\documentclass{ecai}
\usepackage{times}
\usepackage{graphicx}
\usepackage{latexsym}

\usepackage{amsmath,amssymb,amsfonts}
\usepackage{eurosym}
\usepackage{booktabs}
\usepackage[hyphens]{url} 

\usepackage[table]{xcolor}
\usepackage{xspace}
\def\AI{{\rm {\sc AI}}\xspace}
\def\MCTS{{\rm {\sc MCTS}}\xspace}

\def\GGPBASE{{\rm {\sc GGP-BASE}}\xspace}
\def\GGP{{\rm {\sc GGP}}\xspace}
\def\GDL{{\rm {\sc GDL}}\xspace}
\def\GDLII{{\rm {\sc GDL-II}}\xspace}
\def\GDLIII{{\rm {\sc GDL-III}}\xspace}
\def\DLP{{\rm {\sc DLP}}\xspace}
\def\RBG{{\rm {\sc RBG}}\xspace}
\usepackage{booktabs}
\usepackage{amsthm}
\theoremstyle{definition}
\newtheorem{definition}{Definition}
\newtheorem{thm}{Theorem}

\theoremstyle{lemma}
\newtheorem{lemma}{Lemma}

\usepackage{listings}
\begin{document}

\title{Ludii - The Ludemic General Game System}

\author{{\'E}ric Piette \and Dennis J. N. J. Soemers \and Matthew Stephenson \\ \and Chiara F. Sironi \and Mark H. M. Winands \and Cameron Browne\institute{Department of Data Science and Knowledge Engineering, Maastricht University,
the Netherlands, email: \{eric.piette, dennis.soemers, matthew.stephenson, c.sironi, m.winands, cameron.browne\}@maastrichtuniversity.nl}}

\maketitle
\bibliographystyle{ecai}

\begin{abstract}
While current General Game Playing (\GGP) systems facilitate useful research in Artificial Intelligence (\AI) for game-playing, they are often somewhat specialised and computationally inefficient. In this paper, we describe the ``ludemic'' general game system Ludii, which has the potential to provide an efficient tool for \AI researchers as well as game designers, historians, educators and practitioners in related fields. Ludii defines games as structures of ludemes -- high-level, easily understandable game concepts -- which allows for concise and human-understandable game descriptions. We formally describe Ludii and outline its main benefits: generality, extensibility, understandability and efficiency. Experimentally, Ludii outperforms one of the most efficient Game Description Language (\GDL) reasoners, based on a propositional network, in all games available in the Tiltyard \GGP repository. Moreover, Ludii is also competitive in terms of performance with the more recently proposed Regular Boardgames (RBG) system, and has various advantages in qualitative aspects such as generality.
\end{abstract}

\section{INTRODUCTION}
\label{Sec:intro}
The goal of \textit{General Game Playing} (GGP) is to develop artificial agents capable of playing a wide variety of games \cite{pitrat68}. 
Several different software systems for modelling games, commonly called General Game Systems, currently exist for different types of games. This includes deterministic perfect-information games \cite{genesereth05}, combinatorial games \cite{browne09}, puzzle games \cite{shaker13}, strategy games \cite{mahlmann11}, card games \cite{font13} and video games \cite{schaul14,GVGAI19}. 

Since 2005, the \textit{Game Description Language} (\GDL) \cite{love08} has become the standard for academic research in \GGP. 
\GDL is a set of first-order logical clauses describing games in terms of simple instructions. 
While it is designed for deterministic games with perfect information, an extension named ``\GDLII'' \cite{schiffel14} has been developed for games with hidden information, and another extension named ``\GDLIII'' \cite{thielscher17} has been developed for epistemic games.

\subsection{GDL background}

The generality of \GDL provides a high level of algorithmic challenge and has led to important research contributions \cite{bjornsson16} -- especially in {\it Monte Carlo tree search} (\MCTS) enhancements \cite{finnsson08,finnsson10}, with some original algorithms combining constraint programming, \MCTS, and symmetry detection \cite{koriche17}.
Unfortunately, the key structural aspects of games -- such as the board or card deck, and arithmetic operators -- must be defined explicitly from scratch for each game definition. 
\GDL is also limited in terms of potential applications outside of game \AI. 

Game descriptions can be time consuming to write and debug, and difficult to decipher for those unfamiliar with first order logic. The equipment and rules are typically interconnected to such an extent that changing any aspect of the game would require significant code rewriting.
For example, changing the board size from $3$$\times$$3$ to $4$$\times$$4$ in the Tic-Tac-Toe description would require many lines of code to be added or modified.
GDL game descriptions are verbose and difficult for humans to understand, and do not encapsulate the key game-related concepts that human designers typically use when thinking about games. 
Processing such descriptions is also computationally expensive as it requires logic resolution, making the language difficult to integrate with other external applications.
Some complex games can be difficult and time consuming to model (e.g., Go), or are rendered unplayable due to computational costs (e.g., Chess). 
The main GGP/GDL repository \cite{ggpbaserepository} is only extended with a few games every year.

\subsection{Regular Boardgames (RBG)}
\label{Sec:RBG}

In \cite{kowalski19}, an alternative to \GDL called Regular Boardgames (\RBG) is proposed. It comes from an initial work proposed by \cite{Bjornsson12} for using a regular language to encode the movement of pieces for a small subset of chess-like games called Simple Board Games. However, as the allowed expressions are simplistic and applied only to one piece at a time, it cannot express any non-standard behaviour. \RBG extended and updated this idea to be able to describe the full range of deterministic board games. 

The \RBG system uses a low-level language -- which is easy to process -- as an input for programs (agents and game manager), and a high-level language -- which allows for more concise and human readable descriptions. The high-level version can be converted to the low level version in order to provide the two main aspects of a \GGP system: human-readability and efficiency for \AI programs. The technical syntax specification of \RBG can be found in \cite{Kowalski18}. Thanks to this distinction between two languages, it is possible to model complex games (e.g., Amazons, Arimaa or Go), and apply the more common \AI methods (minimax, Monte-Carlo search, reinforcement learning, etc.) to them. Indeed, in the previous languages used for \GGP (including the academic state of the art \GDL), it was difficult to model any complex games, and quite hard to play or reason on any of them in a reasonable amount of time. \RBG has been demonstrated to be universal for the class of finite deterministic games with perfect information, and more efficient than \GDL \cite{kowalski19}. 

\subsection{The Digital Ludeme Project}
\label{Sec:DLP}

The Digital Ludeme Project (\DLP)\footnote{Digital Ludeme Project: \url{www.ludeme.eu}} is a five-year research project, launched in 2018 at Maastricht University, which aims to model the world's traditional strategy games in a single, playable digital database. 
This database will be used to find relationships between games and their components, in order to develop a model for the evolution of games throughout recorded human history and to chart their spread across cultures worldwide. 
This project established a new field of research called \textit{Digital Arch\ae{}oludology} \cite{browne17}.

The \DLP aims to model the 1,000 most influential traditional games throughout history, each of which may have multiple interpretations and require hundreds of variant rule sets to be tested. This is therefore not just a mathematical / computational challenge, but also a logistical one requiring a new kind of General Game System. 
The \DLP deals with traditional games of strategy including most board games, card games, dice games, tile games, etc., and may involve non-deterministic elements of chance or hidden information, as long as strategic play is rewarded over random play; we exclude dexterity games, physical games, video games, etc.

In this paper, we formally introduce Ludii, 
the first complete Ludemic General Game System able to model and play (by a human or AI) the full range of traditonal strategy games. We introduce the notion of ludemes in Section~\ref{Sec:ludemes}, and the ludemic approach that we have implemented in Section~\ref{Sec:ludemebased}. Section~\ref{Sec:ludii} describes the Ludii System itself, its abilities to provide the necessary applications to the Digital Ludeme Project are highlighted in Section~\ref{Sec:benefits}, and the underlying efficiency of the Ludii system in terms of reasoning is demonstrated experimentally in Section~\ref{Sec:xp} by a comparison with one of the best GDL reasoners -- propnets \cite{Sironi17} -- and the interpreter and compiler of the Regular Boardgames (\RBG) system. Finally, Section~\ref{Sec:Conclusion} concludes the paper and describes several future research possibilities.

\section{LUDEMES} 
\label{Sec:ludemes}

The decomposition of games into their component {\it ludemes} \cite{parlett16}, i.e. conceptual units of game-related information, allows us to distinguish between a game's {\it form} (its rules and equipment) and its {\it function} (its emergent behaviour through play). This separation provides a clear genotype/phenotype analogy that makes phylogenetic analysis possible, with ludemes making up the ``DNA'' that defines each game. 

This ludemic model of games was successfully demonstrated in earlier work to evolve new board games from existing ones \cite{browne11}. An important benefit of the ludemic approach is that it encapsulates key game concepts, and gives them meaningful labels. This allows for the automatic description of game rule sets, comparisons between games, and potentially the automated explanation of learnt strategies in human-comprehensible terms. 
Recent work shows how this model can be enhanced for greater generality and extensibility, to allow any ludeme that can be computationally modelled to be defined using a so-called {\it class grammar} approach, which derives the game description language directly from the class hierarchy of the underlying source code library \cite{BrowneB16}. 

This approach provides the potential for a single \texttt{AI} software tool that is able to model, play, and analyse almost any traditional game of strategy as a structure of ludemes. It also provides a mechanism for identifying underlying mathematical correspondences between games, to establish probabilistic relationships between them, in lieu of an actual genetic heritage. 

\section{LUDEMIC APPROACH} 
\label{Sec:ludemebased}

We now outline the {\it ludemic approach} used to model games.

\subsection{Syntax}
\label{Sec:syntax}

\begin{definition}
\label{Def:ludiistate}
A Ludii game state $s$ encodes which player is to move in $s$ (denoted by $mover(s)$), and six vectors each containing data for every possible \textit{location}; $what$, $who$, $count$, $state$, $hidden$, and $playable$. A more precise description of the locations and the specific data in these vectors is given after Definition~\ref{Def:ludiigame}.

\end{definition}

\begin{definition}
\label{Def:ludiisuccessor}
A Ludii successor function is given by
\begin{center}
$\mathcal{T}:(S\setminus S_{ter}, \mathcal{A}) \mapsto S,$
\end{center}
where $S$ is the set of all the Ludii game states, $S_{ter}$ the set of all the terminal states, and $\mathcal{A}$ the set of all possible lists of actions. 
\end{definition}
Given a current state $s \in S \setminus S_{ter}$, and a list of actions $A = \left[ a_i \right] \in \mathcal{A}$, $\mathcal{T}$ computes a successor state $s' \in S$. Intuitively, a complete list of actions $A$ can be understood as a single ``move'' selected by a player, which may have multiple effects on a game state (each implemented by a different primitive action).

\begin{definition}
\label{Def:ludiigame}
A Ludii game is given by a 3-tuple of ludemes $\mathcal{G} = \langle Players, Equipment, Rules \rangle$ where: 
\begin{itemize}

\item $Players = \langle \{ p_0, p_1, \ldots, p_k\}, \mathcal{F} \rangle$ contains a finite set of $k+1$ players, where $k \geq 1$, and a definition of the game's control flow $\mathcal{F} \in \{ Alternating, Simultaneous, Realtime \}$. Random game elements (such as rolling dice, flipping a coin, dealing cards, etc.) are provided by $p_0$, which denotes \textit{nature}. The first player to move in any game is $p_1$, and the current player is referred to as the mover. When $\mathcal{F}$ is omitted, we assume an alternating-move game by default.

\item $Equipment= \langle C^t, C^p \rangle$ where:
\begin{itemize}
\item $C^t$ denotes a list of containers (boards, player's hands, etc.). Every container $c^t = \langle V, E \rangle$, where $c^t \in C^t$, is a graph with vertices $V$ and edges $E$. Every vertex $v_i \in V$ corresponds to a playable site (e.g. a square in Chess, or an intersection in Go), while each edge $e_i \in E$ represents that two sites are adjacent.
   
\item $C^p$ denotes a list of components (pieces, cards, tiles, dice, etc.), some of which may be placed on sites of the containers in $C^t$. We use the convention that the component $c^p_0 \in C^p$ is placed on all ``empty'' sites.
\end{itemize}

\item $Rules$ defines the operations of the game, including:

\begin{itemize}

\item $Start = [a_0, a_1, \ldots, a_k]$ denotes a list of starting actions. The starting actions are sequentially applied to an ``empty'' state (state with $c_0$ on all sites of all containers) to model the initial state $s_0$.

\item $Play : S \mapsto \mathcal{P}(\mathcal{A})$, where $\mathcal{P}(\mathcal{A})$ denotes the powerset of the set $\mathcal{A}$ of all possible lists of legal actions. This is a function that, given a state $s \in S$, returns a set of lists of actions.

\item $End =  (Cond_0(s), \Vec{\mathbb{S}}_0)~\cup~\ldots~\cup~(Cond_e(s), \Vec{\mathbb{S}}_e)$ denotes a set of conditions $Cond_i(s)$ under which a given state $s$ is considered to be terminal. Each termination condition $Cond_i(s)$ leads to a score vector $\Vec{\mathbb{S}}_i$.  

\end{itemize}  

\end{itemize}

\end{definition}

Ludii provides some predefined vectors: \texttt{Win}, \texttt{Loss}, \texttt{Draw}, \texttt{Tie}, and \texttt{Abort}. Moreover, if the \textit{mover} has no legal moves then they are in a (temporary) {\it Stalemate} and must perform the special action \texttt{pass}, unless the {\it End} rules dictate otherwise.
States in which all players were forced to \texttt{pass} for the last complete round of play are abandoned as a \texttt{Draw}.

We specify \textit{locations} $loc = \langle c, v_i, l_i \rangle$ by their container $c = \langle V, E \rangle$, a vertex $v_i \in V$, and a \textit{level} $l_i \geq 0$. Every location specifies a specific site in a specific container at a specific level, where most games only use $l_i = 0$ but stacking games may use more levels. For every such location, a game state $s$ encodes multiple pieces of data, as described in Definition~\ref{Def:ludiistate}. The index of a component located at $loc$ in $s$ is given by $what(s, loc)$, the owner (player index) by $who(s, loc)$, the number of components by $count(s, loc)$, and the internal state of a component (direction, side, promotion status, etc.) by $state(s, loc)$. If the state of a location $loc$ is hidden information for a certain player $p_i$, that is given by $hidden(s, loc, p_i)$. Games in which play is not limited to a board or other graph with a predetermined size, such as \textit{Andantino}, are implemented by pre-allocating a large, invisible graph, and only allowing play to start in the centre of the graph. New locations gradually become playable over time as required by the rules. Whether or not any particular location $loc$ is currently playable in a state $s$ in such games is given by $playable(s, loc)$.

\subsection{Ludii example}
\label{Sec:example}
Following Definition \ref{Def:ludiigame}, multiple Ludemes can be combined together to form complete game descriptions. As an example, we provide below the complete description of the game Tic-Tac-Toe according to the EBNF-style grammar generated by Ludii.

\footnotesize
\begin{lstlisting}
(<@\textbf{game}@> "Tic-Tac-Toe"
  (<@\textbf{players}@> 2)
  (<@\textbf{equipment}@> {
    (board (square 3) (square))
    (piece "Disc" P1) (piece "Cross" P2)
  })
  (<@\textbf{rules}@>
    (<@\textbf{play}@> (to Mover (empty)))
    (<@\textbf{end}@> (line 3) (result Mover Win))
  )
)
\end{lstlisting}
\normalsize

The \texttt{players} ludeme describes the mode of play; a game with alternating turns played between two players. 
The first subset of the \texttt{equipment} ludeme describes the main board as a square $3$$\times$$3$ tiling, with the second subset listing the components as a disc piece for player 1 and a cross piece for player 2. 
Each turn, the mover plays a piece of their colour at any empty cell, which is implemented by \texttt{(to Mover (empty))}. The winning condition for the mover is to create a line of three pieces. Tic-Tac-Toe does not require any $Start$ rules. 

If the board fills before either player wins, then game defaults to a {\it Draw} after both players are forced to pass. 
Note that judicious use of default settings for common game behaviours allows succinct game descriptions.

\section{LUDII SYSTEM}
\label{Sec:ludii}

The next section introduces the Ludii system itself (\url{https://ludii.games/})\footnote{The specific version used to generate results reported in this paper may be downloaded from: \url{https://ludii.games/downloads/Ludii_ECAI2020PaperVersion.jar}} describing both the grammar approach and the core of the system. 

\subsection{Class grammar}
\label{Sec:classgrammar}

Ludii is a complete general game system implemented in Java that uses a {\it class grammar} approach, in which the game description language is automatically generated from the constructors in the class hierarchy of the Ludii source code \cite{BrowneB16}. 
Game descriptions expressed in the grammar are automatically instantiated back into the corresponding library code for compilation, giving a guaranteed 1:1 mapping between the source code and the grammar in using the reflection of Java to instantiate the different constructors to get a Java executable.

Schaul {\it et al.}~\cite{Schaul11} points out that \textit{``any programming language constitutes a game description language, as would a universal Turing machine''}. Ludii effectively makes its programming language (Java) the game description language. 
It can theoretically support any rule, equipment or behaviour that can be programmed in Java. The implementation details are hidden from the user, who only sees the simplified gramma, which summarises the code to be called.

\subsection{The core of Ludii}
\label{Sec:ludiicore}

The core of Ludii is a ludeme library, consisting of a number of classes, each implementing a specific ludeme. A Ludii game $\mathcal{G}$ defining all relevant ludemes (players, equipment, rules) is stored as a single immutable {\tt Game} object. In the context of General Game Playing, displaying any game automatically is important for understanding strategies by \AI players and testing the correctness of game implementations through human play. To this end, all equipment in Ludii implements the {\tt Drawable} interface, which means that each item of equipment is able to draw a default bitmap image for itself at a given resolution, for displaying the board state. Containers $C^t$ are able to draw their current components at the appropriate positions, orientations, states, etc. A {\tt View} object provides the mechanism for showing the current game state on the screen and a {\tt Controller} object provides the mechanism for updating the game state based on user input such as mouse clicks. 
All games available in the system can be played by both humans and/or \AI. 

\begin{figure}[t]
\centering
\includegraphics[scale=0.25]{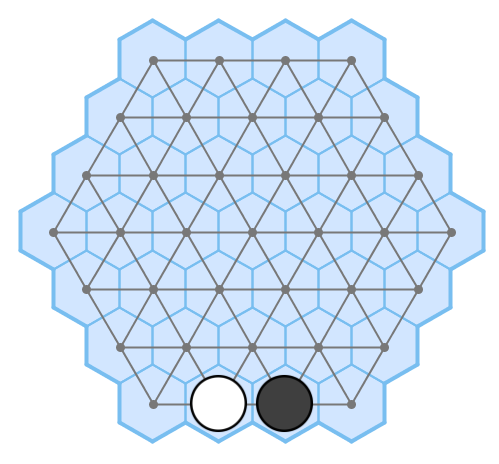}
\caption{A hex board container with hexagonal tiling and the dual of its graph (which is itself a graph).}
\label{fig:LudiiDual}
\end{figure}

As an example, Figure~\ref{fig:LudiiDual} shows a 2-player game $\mathcal{G}$ with $C^t = \{c^t_0\}$, where $c^t_0$ is a hexagonal container with hexagonal tiles. $C^p = \{c^p_0, c^p_1, c^p_2\}$, where $c^p_0$ is the empty component, $c^p_1$ is the white disc for the player $p_1$, and $c^p_2$ the black disc for player $p_2$. The system has a graph representation of the board for visualisation; the vertices, edges, and faces of this graph are depicted in blue. The dual of this graph, which is the graph given by $c^t_0$, is depicted in grey.

The game graph itself can be modified during certain {\it graph games} (e.g. Dots \& Boxes) in which a player's moves involve operations on the graph (e.g. adding or cutting edges or vertices).
Reasoning efficiency can be optimised by pre-generating data such as board corners, exterior vertices, vertices along the top side of the board, etc. within the \texttt{Graph} class, and vertex neighbours indexed by direction, vertices reached by turtle steps, etc. within the \texttt{Vertex} class.

The data vectors $what$, $who$, etc. of a state $s$ are implemented in a collection of {\tt ContainerState} objects. Different representations are implemented in order to minimise the memory footprint and to optimise the time needed to access necessary data for reasoning on any game:

\begin{itemize}

\item Uniform pieces per player (e.g. Tic-Tac-Toe).
\item Distinguished pieces per player (e.g. Chess).
\item Piece state per site (e.g. Reversi).
\item Piece count per site (e.g. Mancala games).
\item Piece stacking (e.g. Lasca).
\item No fixed board (e.g. Dominoes).
\item Hidden information (e.g. Stratego, card games). 

\end{itemize}

Ludii automatically selects the appropriate state type from the rules before creating the {\tt Game} object, to ensure the most suitable representation is used.

Container states are defined using a custom {\tt BitSet} class, called {\tt ChunkSet}, that compresses the required state information into a minimal memory footprint, based on each game's definition. 
For example, if a game involves no more equipment than a board and uniform pieces in $N$ colours, then the game state is described by a {\tt ChunkSet} subdivided into chunks of $B$ bits per board cell, where $B$ is the lowest power of 2 that provides enough bits to represent every possible state per cell (including state 0 for the empty cells).\footnote{Chunk sizes are set to the lowest power of 2 to avoid issues with chunks straddling consecutive {\tt long} values.}

Using the game state and the different ludemes describing the game rules, the system can compute the legal moves for any state. The tree of ludemes is evaluated to return the list of {\tt Action} objects applicable for the mover. 
Each {\tt Action} object describes one or more atomic actions to be applied to the game state to execute a move. 
Actions typically include adding or removing components to/from containers, or changing component counts or states within containers. 

\begin{definition}
\label{def:trial}
A trial $\tau$ is a sequence of states $s_i$ and action lists $A_i$: 
\begin{center}
$\langle s_0, A_1, s_1, \ldots, s_{f-1}, A_f, s_f \rangle$
\end{center}
such that $f \geq 0$, and for all $i \in \{1, \ldots f\}$,
\begin{itemize}
    \item the played action list $A_i$ is legal for the $mover(s_{i-1})$
    \item states are updated: $s_i = \mathcal{T}(s_{i-1}, A_{i})$
    \item only $s_f$ may be terminal: $\{s_0, \ldots, s_{f-1}\} \cap S_{ter} = \emptyset$
\end{itemize}
\end{definition}

$\tau$ is stored in a {\tt Trial} object, providing a complete record of a game played from start to end, including the moves made.

Any reasoning on any game can be parallelised using separate trials per thread. 
All the data members of the {\tt Game} object are constant and can therefore be shared between threads. 
A thread will be able to use a {\tt Trial} object to compute any playouts from any state. 
On the system each {\tt AI} object describes the AI implementation chosen for each player, including computational budget/time limits, hints such as features for biasing playouts \cite{browne19}, etc.

\section{BENEFITS AND KEY PROPERTIES}
\label{Sec:benefits}

Ludii is being designed and implemented primarily to provide answers to the questions raised by the \DLP, but will stand alone as a platform for general games research in areas including \AI, design, history and education. 
Ludii provides many advantages over existing \GGP systems, as follows:  

\begin{figure}[t]
\centering
   \begin{minipage}[c]{.48\linewidth}
   \scriptsize
\fbox{%
  \parbox{.96\columnwidth}{%
  	\texttt{%
  		(chessBoard (square 8))  \\
  	}%
  }%
}
   \end{minipage} \hfill
         \begin{minipage}[c]{.48\linewidth}
   \scriptsize
\fbox{%
  \parbox{.97\columnwidth}{%
  	\texttt{%
  		 (chessBoard (hexagonal 5)) \\
  	}%
  }%
}
   \end{minipage} \hfill
   ~ \\
   
      \begin{minipage}[c]{.47\linewidth}
\centering
\includegraphics[width=.98\columnwidth]{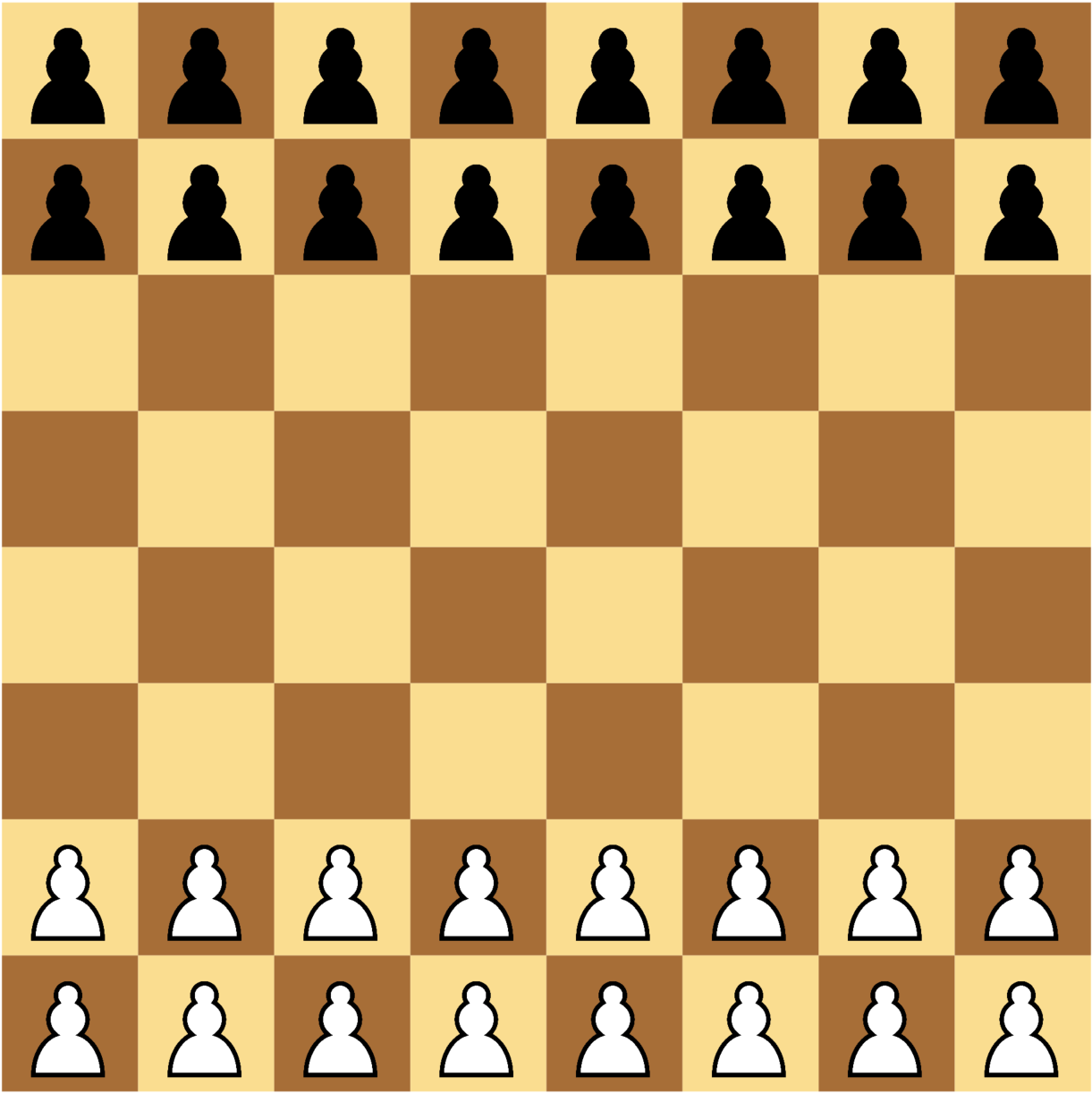}
   \end{minipage} \hfill
      \begin{minipage}[c]{.47\linewidth}
\centering
\includegraphics[scale=0.2]{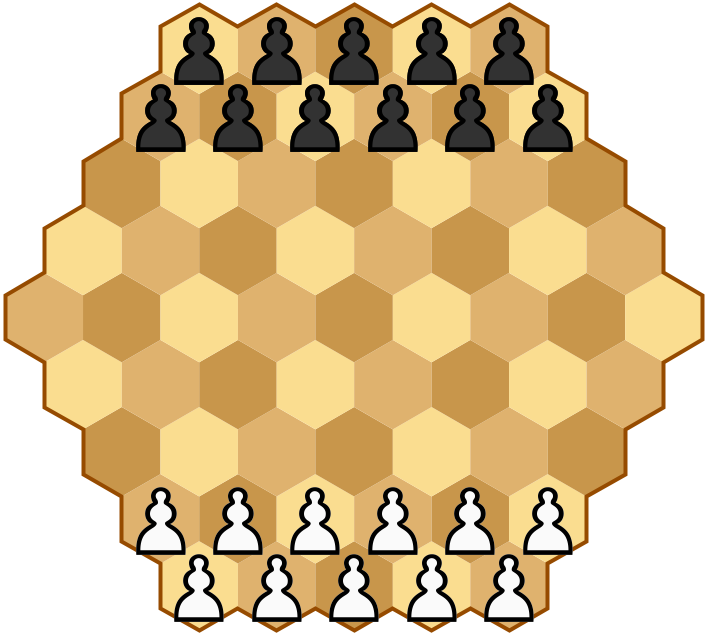}
   \end{minipage} \hfill
\caption{The game of Breakthrough on square (left) and hexagonal (right) board tilings.}
\label{Fig:Ludiibreakthrough}
\end{figure}

\paragraph{Simplicity:} Simplicity refers to the ease with which game descriptions can be created and modified, and can be estimated by the number of tokens required to define games. This data is shown for different games in GDL, RBG and Ludii in Table \ref{table:results1}.  
Describing a game with the ludemic approach is typically much simpler compared to a logic-based approach (e.g. Ludii requires only $24$ tokens for Tic-Tac-Toe and $548$ for Chess, whereas \GDL requires $381$ and $4,392$ tokens respectively).
Ludemic game descriptions can also be easily modified to test different sizes, geometries or rules. For example, changing the size or shape of a board (e.g. Figure~\ref{Fig:Ludiibreakthrough}) can be accomplished by modifying a single parameter, while the same change in \GDL requires many lines of code to be added or modified.

Note that, in the case of Ludii, these results often include additional tokens to provide customisation options, essentially encoding multiple variants of games within a single file. For example, Ludii includes options for many different board sizes (ranging from $3$$\times$$3$ to $19$$\times$$19$), as well as an option for an inversed ``mis{\`e}re'' win condition, all in the same \textit{Hex} game description file with 129 tokens. \GDL and RBG both require completely new files for such modifications, and their game descriptions tend to grow quickly for larger board sizes.

{
\centering
\begin{table}
\caption{A comparison of the number of tokens to describe games used between GDL, RBG and Ludii.}
\begin{center}
\footnotesize
\begin{tabular}{@{}lrrr@{}}
\toprule
\bf Game &\bf {\tt \bf GDL} & \bf {\bf RBG} & \bf {\tt \bf Ludii} \\
\midrule
Amazons ($10$$\times$$10$)  & 1,158 & 204 & 51 \\
Breakthrough ($8$$\times$$8$)  & 670 & 134 & 72 \\
Chess ($8$$\times$$8$) & 4,392 & 641 & 548 \\
Chinese Checkers & 1530 & 418 & 243 \\
Connect 4 ($6$$\times$$7$)  & 751 & 155 & 29 \\
Dots $\&$ Boxes ($6$$\times$$6$)  & 689 & $\times$ & 131 \\
English Draughts  & 1,282 & 263 & 161 \\
Gomoku ($15$$\times$$15$)  & 514 & 324 & 32 \\
Hex ($9$$\times$$9$)  & 702 & 198 & 129 \\
Hex ($11$$\times$$11$)  & $\times$ & 245 & 129 \\
Reversi ($8$$\times$$8$)  & 894 & 311 & 78 \\
Tic-Tac-Toe ($3$$\times$$3$) & 381 & 101 & 24 \\
Tron ($10$$\times$$10$) & 405 & $\times$ & 50 \\
Wolf $\&$ Sheep ($8$$\times$$8$) & 794 & $\times$ & 55 \\
\bottomrule
\end{tabular}
\normalsize
\end{center} 
\label{table:results1}
\end{table}
}

\paragraph{Clarity:} Clarity refers to the degree to which game descriptions would be self-explanatory to non-specialist readers. 
The logic-based game descriptions of \GDL are often difficult for humans to interpret.
Appropriately-named macros in RBG can improve readability, but the game descriptions still require knowledge of regular languages to fully understand. In Ludii, the Java classes that define each ludeme are named using meaningful English labels, providing convenient definitions for the concepts involved. This is especially useful for games that encapsulate more complex mathematical concepts (geometry, algebra, arithmetic, etc.) within their component ludemes.

\paragraph{Generality:} Generality refers to the scope of games covered by the system without the need for extensions.. 
As Ludii uses the class grammar approach to describe the ludemes, it can theoretically support any game that can be programmed in Java, The version of Ludii described in this paper already includes many different game types beyond those that can be implemented in \GDL and RBG. This includes stochastic games, games with hidden information, simultaneous-move games, and one proof-of-concept real-time game. The current version of RBG is restricted to deterministic, perfect-information, alternating-move games. GDL (without extensions) is restricted to deterministic, perfect-information games.

\paragraph{Extensibility:} Extensibility refers to the ease with which new functionality can be added to the system. 
The current version of Ludii provides approximately 100 games, and already contains most of the systems that are expected to be necessary to support most classes of games (like stochastic games, hidden information games, etc.).
Extending Ludii simply involves adding new classes to the ludeme library, which are then automatically subsumed into the grammar, making extensibility very open-ended. 
Extending \GDL involves significant modifications to the core model and program. Similarly, RBG is expected to require significant modifications to its language and underlying systems to support new classes of games, such as stochastic or hidden information games.

\paragraph{Efficiency:} Since the Ludii programmer has complete control of the underlying code -- within the constraints of the API and programming guidelines -- it is possible to optimise ludemes at any desired level. 
There is of course a trade-off between game optimisation and description detail. The more optimised a game is, the shorter its description tends to be and the less detail we know about it. 
This has profound implications for the \DLP, in which the ability to reliably compare games for similarity is a key requirement. In Ludii, we sometimes prioritise aspects such as clarity or generality over efficiency. Section \ref{Sec:results} describes different experiments ran in order to compare Ludii to \GDL and \RBG in terms of efficiency.

\paragraph{Evolvability:} This refers to the likelihood that randomly evolving game descriptions will produce viable children that resemble their parents. \GDL game descriptions tend to involve complex chains of logical operations that must be crafted with great care. Randomly applying crossovers and mutations between \GDL descriptions is extremely unlikely to produce playable results, let alone improve on the parents.
Conversely, the ludemic approach is ideally suited to evolutionary approaches such as {\it genetic programming}~\cite{Koza1992} and has already proven successful in evolving new high quality games~\cite{browne09}.

\paragraph{Cultural Applications:} Aside from its \GGP benefits, Ludii also has several applications as a tool for the new domain of Digital Arch\ae{}oludology \cite{browne17}. 
The Ludii system is linked to a server and database that stores relevant cultural and historical information about the games. This information will not only provide additional real-world context, but will allow us to reconstruct viable and historically authentic rule-sets for games with incomplete information, develop a ``family tree'' of traditional games, and help map the spread of games throughout history. 

\paragraph{Universality:} While Ludii supports a wide range of games, including nondeterministic and hidden information games, we cannot prove the universality of its full grammar within the scope of this paper. 
We instead show that Ludii is universal for finite deterministic perfect information games:

\begin{thm}
\label{theorem1}
Ludii is universal for the class of finite deterministic games with perfect information.
\end{thm}

Similar to \cite{kowalski19}, we formalise a finite, deterministic, $k$-player game with perfect information as a tuple $(k, T, \iota, \upsilon)$, where:
\begin{itemize}
    \item $k \in \mathbb{N}$ indicates the number of players.
\item  $T$ is a finite tree with:
\begin{itemize}
    \item Nodes $S$ (also referred to as game states).
    \item An initial state $s_0 \in S$ (the root node of $T$).
    \item Terminal states $S_{ter} \subseteq S$ (leaf nodes of $T$).
    \item A predecessor function $f:(S \setminus \{ s_0 \}) \mapsto S$, such that $f(s)$ denotes the parent of $s$ in $T$.
\end{itemize}
\item $\iota : (S \setminus S_{ter}) \mapsto \{0, \ldots, k\}$ indicating which player has the control in a given state.
\item $\upsilon: S_{ter} \mapsto \mathbb{R}^k$, such that $\upsilon(s)$ denotes the vector of payoffs for $k$ players for any terminal state $s \in S_{ter}$.
\end{itemize}
This is equivalent to the formalisation of $k$-player extensive form games by \cite{rasmusen07}, excluding elements required only for non-determinism or imperfect information.

We prove that, given any arbitrary finite, deterministic game with perfect information as defined above, a Ludii game can be constructed such that there is a one-to-one mapping between states and state transitions between the original game and the Ludii game. The intuition of our proof is to construct a Ludii game where the game board is represented by a graph with an identical structure to the full game tree $T$. The Ludii game is played by moving a single token, placed on the ``root node'' in the initial game state, along the graph until a leaf node is reached. For any state $z$ in the original game, there is a corresponding state $s$ in the Ludii game such that the token is located on the vertex corresponding to the position of $z$ in $T$. Note that explicitly enumerating the complete game tree as a graph is unlikely to be the most optimal representation for most games, but it demonstrates that Ludii is capable of representing all such games.

\begin{definition}
\label{Def:ConstructedLudiiGame}
Let $\mathcal{D} = (k, T, \iota, \upsilon)$ denote a finite, deterministic, $k$-player game with perfect information as formalised above. We define a related Ludii game $\mathcal{G}(\mathcal{D}) = \langle Players, Equipment, Rules \rangle$, where $Rules = \langle Start, Play, End \rangle$, such that:
\begin{itemize}
    \item $Players = \langle \{p_0, p_1, \ldots, p_k\}, Alternating\rangle$, where all $p_i$ for $i \geq 1$ correspond to the $k$ different players. The nature player $p_0$ will remain unused in deterministic games.
    
    \item $Equipment = \langle \{ c^t_0 \}, \{ c^p_0, c^p_1 \} \rangle$. The only container $c^t_0 = \langle V, E \rangle$ is a graph with a structure identical to the tree $T$ of the original game $\mathcal{D}$. Due to the structure of $c^t_0$ being identical to the structure of $T$, we can uniquely identify a vertex $v(z)$ for any state $z \in T$ from the original game. For any such vertex -- except for $v(s_0)$ -- we can also uniquely identify an adjacent ``parent'' vertex $p(v(z))$, such that $p(v(z)) = v(f(z))$; the parent of a vertex corresponds to the predecessor of the corresponding state in $T$.
    
    \item The $Start$ rules are given by a list containing only a single action. This action creates the initial game state by placing the $c^p_1$ token on the site $v(s_0)$ of $c^t_0$ that corresponds to the root node of $T$.
    
    \item Let $s$ denote any non-terminal Ludii game state, such that there is exactly one site $v(z)$ for which $what(s, \langle c_0^t, v(z), 0 \rangle) = c^p_1$. Let $z$ denote the state in the original game that corresponds to the site $v(z)$. Let $g(z)$ denote the children of $z$ in $T$. Given $s$, we define $Play(s)$ to return a set $\{ A_i \}$ of lists of actions $A_i$, with one list of actions for every child node $z' \in g(z)$. Each of those lists contains two primitive actions; one that takes the token $c^p_1$ away from $v(z)$ (replacing it with the ``empty'' token $c^p_0$), and a second action that places a new token $c^p_1$ on the site $v(z')$ of $c^t_0$ that corresponds to the child $z' \in T$.
    
    \item The end rules are given by $End =  \{ (Cond_i(\cdot), \Vec{\mathbb{S}}_i) \}$. For any terminal game state $z_i \in S_{ter}$, let $v(z_i)$ denote the site in the graph $c^t_0$ that corresponds to the position of $z_i$ in $T$. We add a tuple $(Cond_i(\cdot), \Vec{\mathbb{S}}_i)$ to $End$ such that $Cond_i(s)$ returns true if and only if $what(s, loc) = c^p_1$ for $loc = \langle c^t_0, v, 0 \rangle$, and $\Vec{\mathbb{S}}_i = \upsilon(z_i)$. Intuitively, we use a separate end condition for every possible terminal state $z_i \in S_{ter}$ in the original game $\mathcal{D}$, which checks specifically for that state by making sure the $c^p_1$ token is placed on the matching vertex $v(z_i)$.
    
    \item Let $s$ denote any non-terminal Ludii game state, such that there is exactly one site $v(z)$ for which $what(s, \langle c_0^t, v(z), 0 \rangle) = c^p_1$. Let $z$ denote the state in the original game that corresponds to the site $v(z)$. Then, we define $mover(s) = \iota(z)$.
\end{itemize}
\end{definition}

\begin{lemma}
\label{Lemma:EveryStateOneToken} Let $\mathcal{G}(\mathcal{D})$ denote a Ludii game constructed as in Definition~\ref{Def:ConstructedLudiiGame}. Every game state $s$ that can be reached through legal gameplay in such a game has exactly one vertex $v \in c^t_0$ such that $what(s, \langle c_0^t, v, 0 \rangle) = c^p_1$, and $what(s, \langle c_0^t, u, 0 \rangle) = c^p_0$ for all other vertices $u \neq v$.
\end{lemma}
Intuitively, this lemma states that every game state reachable through legal gameplay has the $c^p_1$ token located on exactly one vertex, and that all other vertices are always empty (indicated by $c^p_0$).

\begin{proof}
Let $s_0$ denote the initial game state. The $Start$ rules are defined to place a single $c^p_1$ token on $v(z_0)$, where $z_0$ denotes the initial state in the $\mathcal{D}$ game, which means that the lemma holds for $s_0$.

Let $s$ denote any non-terminal game state for which the lemma holds. Then, the assumptions in Definition~\ref{Def:ConstructedLudiiGame} for an adequate definition of $Play(s)$ are satisfied, which means that $\{ A_i \} = Play(s)$ is a non-empty set of lists of actions, one of which must be selected by $mover(s)$. Every $A_i$ is defined to take away the token $c^p_1$ from the vertex it is currently at, and to place it on exactly one new vertex. This means that the lemma also holds for any successor state $\mathcal{T}(s, A_i)$, which proves the lemma by induction.
\end{proof}

We are now ready to prove Theorem \ref{theorem1}:

\begin{proof}
Let $\mathcal{D}$ denote any arbitrary game as formalised above, with a tree $T$. Let $\mathcal{G}(\mathcal{D})$ denote a Ludii game constructed as described in Definition~\ref{Def:ConstructedLudiiGame}. We demonstrate that for any arbitrary traversal through $T$, from $s_0$ to some terminal state $z_{ter} \in S_{ter}$, there exists an equivalent trial $\tau$, as defined in Definition 4, in $\mathcal{G}(\mathcal{D})$. By ``equivalent'' trial, we mean that the sequence of states traversed is equally long, the order in which players are in control is equal, and the payoff vectors at the end are equal.

Let $z_0, z_1, \dots, z_f$ denote any arbitrary line of play in the original game $\mathcal{D}$, such that $z_0$ is the initial game state, and $z_f \in S_{ter}$. By construction, the initial game state $s_0$ of $\mathcal{G}(\mathcal{D})$ has the token $c^p_1$ placed on the vertex $v(z_0)$ corresponding to the root node of $T$. This means that we have a one-to-one mapping from $z_0$ to $s_0$, where $what(s_0, \langle c_0^t, v(z_0), 0 \rangle) = c^p_1$.

Let $z_i$ denote some non-terminal state in the sequence $z_0, z_1, \dots, z_{f-1}$, such that we already have uniquely mapped $z_i$ to a Ludii state $s_i$ where $what(s_i, \langle c_0^t, v(z_i), 0 \rangle) = c^p_1$. Lemma~\ref{Lemma:EveryStateOneToken} guarantees that the assumptions required for Definition~\ref{Def:ConstructedLudiiGame} to adequately define $Play(s_i)$ are satisfied. Furthermore, we know that $\iota(z_i) = mover(s_i)$, which means that the same player is in control. The definition of $Play(s_i)$ ensures that there is exactly one legal list of actions $A_i$ such that $\mathcal{T}(s_i, A_i) = s_{i+1}$, where $what(s_{i+1}, \langle c_0^t, v(z_{i+1}), 0 \rangle) = c^p_1$ (note that $z_{i+1}$ must be a successor of $z_i$ in $T$). We pick this $s_{i+1}$ to uniquely map to $z_{i+1}$. 

By induction, this completes the unique mapping between sequences of states $z_0, z_1, \dots, z_f$ and $s_0, s_1, \dots, s_f$, uniquely specifies the lists of actions $A_i$ that must be selected along the way, and ensures that $z_i$ is always mapped to a state $s_i$ such that the $c^p_1$ token is placed on $v(z_i)$. This last observation ensures that one of the $End$ conditions in $\mathcal{G}(\mathcal{D})$ triggers for $s_f$, and that the correct payoff vector $\Vec{\mathbb{S}} = \upsilon(z_f)$ is selected.
\end{proof}

\section{EXPERIMENTS}
\label{Sec:xp}

{
\centering
\begin{table*}
\caption{An experimental comparison of the number of playouts per second between GDL, RBG, and Ludii.}
\label{table:results2}
   \begin{center}
\footnotesize
\begin{tabular}{@{}l|rrrrrrr@{}}
\toprule
\bf  & \bf {\tt \bf } & \bf {\tt \bf RBG} & \bf {\tt \bf RBG} &\bf {\tt \bf } & \bf {\bf Rate} & \bf {\bf Rate RBG} & \bf {\bf Rate RBG} \\
\bf Game & \bf {\tt \bf GDL} & \bf {\tt \bf Interpreter} & \bf {\tt \bf Compiler} &\bf {\tt \bf Ludii} &\bf {\tt \bf GDL} & \bf {\bf Interpreter} & \bf {\bf Compiler}\\
\midrule
Amazons & 185 & 1,911 & 7,883 & 3,916 & 21.17 & 2.05 & 0.50\\
Breakthrough & 1,123 & 5,345 & 17,338 & 3,795 & 3.38 & 0.71 & 0.22 \\
Chess & \textit{0.06} & 81 & 704 & 14 & 233.33 & 0.17 & 0.02 \\
Chinese Checkers & 297 & 236 & 1,145 & 780 & 2.62 & 3.31 & 0.68\\
Connect-4 & 13,664 & 41,684 & 174,034 & 78,925 & 5.78 & 1.89 & 0.45\\
Dots $\&$ Boxes & 1672 & $\times$ & $\times$ & 5,671 & 3.39 & $\times$ & $\times$\\
English Draughts & 872 & 3,130 & 14,781 & 8,052 & 9.23 & 2.57 & 0.54 \\
Gomoku & 927 & 1,338 & 2,405 & 36,445 & 39.31 & 27.24 & 15.15 \\
Hex ($9$$\times$$9$) & 195 & 5,669 & 13,360 & 21,987 & 112.75 & 3.88 & 1.65\\
Hex ($11$$\times$$11$) & $\times$ & 2,757 & 6,282 & 12,303 & $\times$ & 4.46 & 1.96\\
Reversi & 203 & 1,349 & 7,434 & 1,438 & 7.08 & 1.07 & 0.19\\
Tic-Tac-Toe & 85,319 & 199,823 & 473,372 & 545,567 & 6.39 & 2.73 & 1.15\\
Tron & 121,988 & $\times$ & $\times$ & 238,129 & 1.95 & $\times$ & $\times$\\
Wolf $\&$ Sheep & 5,533 & $\times$ & $\times$ & 23,106 & 4.18 & $\times$ & $\times$\\
\bottomrule
\end{tabular}
\normalsize
\end{center} 
\end{table*}
}

The Ludii System uses \MCTS as its core method for AI move planning, which has proven to be a superior approach for general games in the absence of domain specific knowledge \cite{finnsson10}. 
\MCTS playouts require fast reasoning engines to achieve the desired number of simulations. 
Hence, we use {\it flat Monte Carlo} playouts (i.e., trials $\tau$ where $s_f \in S_{ter}$) 
as the metric for comparing the efficiency of Ludii to other GGP systems. 

\subsection{Experimental design}
\label{Sec:xpdesign}

In the following comparison, we compare \GDL, \RBG and Ludii based on the number of random playouts from the initial game state obtained per second. 

For \GDL, we used the fastest available game implementation from \cite{ggpbaserepository} and tested one of the most efficient reasoners based on {\it propositional networks} or ``propnets''~\cite{Sironi17}, implemented in \GGPBASE\footnote{\url{https://github.com/ggp-org/ggp-base}}.
Propnets speed up the reasoning process with respect to custom made or Prolog-based reasoners by translating the \GDL rules into a directed graph that resembles a logic circuit, whose nodes correspond to either logic gates or \GDL propositions that represent the state, players' moves and other aspects of the game.
Information about the current state can be computed by setting the truth value of the propositions that correspond to the state and propagating these values through the graph. 
Setting and propagating the truth values of the propositions that correspond to the players' actions allows us to compute the next state. \RBG provides an interpreter and a compiler to perform reasoning, both of which we compare to Ludii.

Every experiment was conducted on a single core of an Intel(R) Core(TM) i7-8650U CPU at @ 1.90 GHz, 2.11 GHz, running for at least ten minutes, allowing at most 4GB RAM to be used.

\subsection{Results}
\label{Sec:results}

The results of our experiments for a selection of games available in \GDL and/or \RBG, are shown in Table \ref{table:results2}, showing the number of random playouts obtained per second. All included games have, to the best of our knowledge, complete and correct game rules in Ludii. 

Table \ref{table:results3} highlights our results for a selection of games, including several historical games, that have no \GDL or \RBG equivalent. 
The fact that no existing GGP system supported the full range of games required for the \DLP was a driving motivation in developing Ludii.

{
\centering
\begin{table}
\caption{The average number of playouts per second for games unavailable in \GDL or \RBG.}
\begin{center}
\footnotesize
\begin{tabular}{@{}lr@{}||@{}lr@{}}
\toprule
\bf Game &\bf {\tt \bf Ludii~~~} & \bf ~~~~Game &\bf {\tt \bf Ludii} \\
\midrule
Alquerque ($5$$\times$$5$) & 12,283~~~ & ~~~ Oware  & 9,622 \\
Connect-6 ($19$$\times$$19$) & 14,126~~~ & ~~~  Royal Game of Ur  & 1,537  \\
Dara & 2,141~~~ & ~~~ Senet ($3$$\times$$10$)  & 664  \\
Fanorona & 2,952~~~ & ~~~ Stratego & 108 \\
Hnefatafl ($11$$\times$$11$) & 197~~~ & ~~~ Surakarta ($6$$\times$$6$) & 672 \\
Konane & 4,653~~~ & ~~~ Tant Fant  & 43,129 \\
Mu Torere & 3,438~~~ & ~~~ Yavalath  & 175,525 \\
\bottomrule
\end{tabular}
\normalsize
\end{center} 
 \label{table:results3}
\end{table}
}

\subsection{Discussion}
\label{Sec:discussion}

Ludii outperforms \GDL in terms of efficiency for all games tested. 
Ludii is at least two times faster for all but one game.
For simpler games, board size is highly correlated with speed improvement; Ludii is more than six times faster for the standard $3$$\times$$3$ Tic-Tac-Toe, but almost forty times faster for the Gomoku $15$$\times$$15$.  
For more complex games -- such as Amazons and Hex -- Ludii is once again more efficient than \GDL (over twenty times faster in these cases).

The greatest speed disparity is for Chess, with a rate greater than $200$.
The \GDL description of Chess cannot be translated to a propnet because its size exceeds the memory, therefore we had to use the \GDL Prover from the \GGPBASE for comparison. The \GGPBASE Prover is generally slower for complex games with respect to the propnet, explaining the low number of playouts for Chess ($0.06$). 

In most games, Ludii outperforms the interpreter of \RBG, but is outperformed by the compiler of \RBG (with playout counts typically still being within the same order of magnitude). In games like Hex and Gomoku, Ludii still outperforms also the compiler of RBG. We note that the RBG compiler requires a significant amount of initialisation time for some games, with initialisation times of over 10 seconds being reported for some individual games \cite{Kowalski18}. In contrast, Ludii's unit test which compiles \textit{all} included games (over 100) takes less than 10 seconds in total. Some of the largest differences in performance in comparison to RBG, such as the difference in Chess, are due to deliberate trade-offs in favor of aspects like clarity and evolvability.

\section{CONCLUSION}
\label{Sec:Conclusion}

The proposed ludemic General Game System Ludii outperforms \GDL~ -- the current standard for academic \AI research into GGP -- in terms of reasoning efficiency, and is competitive with RBG. It also has advantages in terms of simplicity, clarity, generality, extensibility and evolvability, and represents a significant step forward for general game playing research and development.

The potential benefits of this new \GGP approach present several opportunities for future \AI work. For example, features discovered by reinforcement learning could be automatically visualised for any game to possibly reveal useful strategies relevant to that game, or provided as human-understandable descriptions based on ludemes with meaningful plain English labels. 
Another work in progress includes improving \AI playing strength by biasing \MCTS with features automatically learnt through self-play.

Ludii is the first general game system that can model the complete scope of games needed for the \DLP. Most of the major game types have already been implemented within the Ludii system, providing valuable resources for developing and evaluating new game AI techniques. Ludii also has applications beyond this, allowing researchers in fields such as history, game design, mathematics and education to analyse the inherent similarities and strategic properties within a wide range of traditional strategy games.

\ack This research is part of the European Research Council-funded Digital Ludeme Project (ERC Consolidator Grant \#771292). 
This work is partially supported by The Netherlands Organisation for Scientific Research (NWO) in the framework of the GoGeneral Project (Grant No. 621.001.121).

\bibliography{paper}
\end{document}